# HAD-GAN: A Human-perception Auxiliary Defense GAN to Defend Adversarial Examples

Wanting Yu, Hongyi Yu*, Lingyun Jiang, Mengli Zhang,Kai Qiao
Information Engineering University, Zhengzhou 450001, China
*ywan1107@163.com, xxgcmaxyu@163.com, yunlord@outlook.com, zml1122y@163.com, qiaokai1992@gmail.com*

**Corresponding author: Hongyi Yu*

## Abstract

*Adversarial examples reveal the vulnerability and unexplained nature of neural networks. Studying the defense of adversarial examples is of considerable practical importance. Most adversarial examples that misclassify networks are often undetectable by humans. In this paper, we propose a defense model to train the classifier into a human-perception classification model with shape preference. The proposed model comprising a texture transfer network (TTN) and an auxiliary defense generative adversarial networks (GAN) is called Human-perception Auxiliary Defense GAN (HAD-GAN). The TTN is used to extend the texture samples of a clean image and helps classifiers focus on its shape. GAN is utilized to form a training framework for the model and generate the necessary images. A series of experiments conducted on MNIST, Fashion-MNIST and CIFAR10 show that the proposed model outperforms the state-of-the-art defense methods for network robustness. The model also demonstrates a significant improvement on defense capability of adversarial examples.*

## 1. Introduction

Deep learning (DL) (LeCun et al., 2015) has demonstrated advancements in various tasks of artificial intelligence, including image recognition (He et al., 2016), natural language processing (Collobert & Weston, 2008), and speech processing (Hinton et al., 2016). These advancements are attributed to deep neural networks (DNNs), which can represent complex probability distribution over high dimensional data. Despite these advancements, DNNs remain imperfect. In particular, these networks show weaknesses considering adversarial examples when compared with humans (Akhtar & Weston, 2018). Adversarial examples can effectively fool a neural network to change its predictions, and the human eye cannot distinguish such examples from the original images. Moreover, adversarial examples can be transferred across different models (Akhtar & Mian, 2018). Improving the robustness of neural networks and finding effective defenses against these attacks are important, especially for security-critical applications.

In recent years, many defense methods against adversarial examples have been proposed, and these methods can be divided into two categories. The first category is enhancement of the network robustness. Adversarial training (Goodfellow et al., 2014; Araujo et al., 2019) and defensive distillation (Papernot et al., 2015) for classifiers are two typical and excellent methods. These defense methods often require attack patterns to be known a priori and are satisfactory against white-box attacks but poor for black-box attacks. The second category is a variety of data preprocessing methods. For example, PixelDefend (Song et al., 2018), high-level representation guided denoiser (HGD) (Liao et al., 2018), and ComDefend (Jia et al., 2019) are designed to combat against adversarial examples. Detection-based defense against adversarial examples from the steganalysis point of view (Liu et al., 2019) also shows excellent performance advantages. Implementing the latter category of the method is easier because retraining the neural network is no longer required. This type of preprocessing method mostly treats adversarial examples defense as a denoising process and does not fully explain the effect of adversarial examples on the network. Research on improving the robustness of the network itself is limited.

Through the complete theoretical derivation and experimental verification, researchers from MIT first proposed that "adversarial examples are not bugs, they are features" (Ilyas et al., 2019). By extracting and analyzing the robustness and non-robustness of the image, they indicated that the neural network does not necessarily identify the image in the way of human decision-making but may instead capture some pixel information that does not affect humans. Then, they studied the dataset obtained by robust training and found that the features with robust training are more "human-aligned" (Engstrom et al., 2019). Their research indicated a new research direction for studying adversarial examples for the first time and provided a complete theoretical framework. The research results also further confirm our idea. However, the robust and non-robust features in their paper are mainly extracted by pre-trained models. Their experiments did not provide a concrete method that could use human perception to improve the model robustness effectively against adversarial examples.

Simultaneously, the samples encountered in human learning are rich and diverse with interference and imagination. One of the most promising methods for unsupervised learning in complex distribution in recent years, generative adversarial networks (GANs) (Goodfellow et al., 2014), have the same effect as the human imagination. Originally introduced by Goodfellow et al. (2014), the GAN is a novel and successful generation model and deep learning framework. In this paper, the excellent image generation capability of GAN is used to improve the pre-training network structure, and an auxiliary defense GAN model close to human experience is designed.

Inspired by human thinking and learning growth, this paper introduces the rich imagination of people and the human eye preference for shape into the design of the network structure and provides another "human" intelligence to the neural network. Specifically, the idea of designing a network has the following three aspects:

- When the human eye observes the target, the attention of the target shape is larger than the texture. By observing the confrontation sample, confusing the classification network but not deceiving the human eye is possible. The shape characteristics of the original target may maintain the macro shape.
- The existing data used to train the classification network are usually clear and clean (no noise). However, from the perspective of human learning, the information obtained is usually varied and ambiguous. The learning of such rich information also strengthens the capability of the human eye to resist noise.
- Human learning is not only based on existing limited samples but also from samples generated by human imagination. This process is similar to the training process of GAN. When designing the structure, we also attempt to control the loss function; thus, the training is generated in a specific direction.

Our contributions are as follows.

1. We propose HAD-GAN, a novel model that can introduce human shape perception preferences into adversarial examples defense by adding a texture transfer module. Our defense mechanism changes the training of the target classifier into a dynamic learning process similar to humans.
2. For the first time, we associate GAN with human imagination. Our model links the last layer of the target classifier in parallel with a discrimination layer and completes the defense training only during the GAN training process.
3. At present, many defense methods are conducted under conditions known in the attack method. The proposed model does not need to assume that the attack method is known a priori and is designed to improve the robustness of the network itself.
4. Defense tests were performed on MNIST, Fashion-MNIST, and CIFAR10 datasets at different target networks and attack methods. We also experimentally verified our defense model. Simultaneously, our model achieves better performance than that of the current defense methods (adversarial training, defensive distillation, Defense-GAN and BCGAN).

The remainder of the paper is organized as follows. The necessary related work regarding existing attack and defense methods, GANs, and shape versus texture property studies of neural networks is presented in Section 2. Our defense model, which is called HAD-GAN, is formally motivated and introduced in Section 3. Then, the experimental settings and results are presented and comprehensively described in Section 4. Finally, the conclusion is given in Section 5.

# 2. Related work

## 2.1 Existing attack and defense methods

### 2.1.1 Attack methods

Finding a good attack is a bi-criteria optimization problem, in which the attacker attempts to minimize the norm of the perturbation while trying to maximize the loss function. Various attack models and algorithms have been used to target classifiers. The present common attacks are mainly divided into white- and black-box attacks. In black-box attack, the attacker has no access to the parameters and training data set of the classification model. In the white-box attack, the attacker can obtain all the parameters of the classification model, such as network structure and permissions and details of the defense mechanism.

**Fast Gradient Sign Method (FGSM)** (Goodfellow et al., 2015) Given an image $x$ and its corresponding true label $y$, the FGSM attack sets the perturbation $\delta$ to

$$\delta = \epsilon \cdot \mathrm{sign}(\nabla_x J(x, y)). \tag{1}$$

FGSM (Goodfellow et al., 2015) was designed to be extremely fast rather than optimal. The method simply uses the gradient sign at every pixel to determine the direction in which to change the corresponding pixel value.

**Projected Gradient Descent (PGD) attack** (Sinha et al., 2018) PGD is an iterative attack method proposed by Madry et al. in 2017 and can be regarded as a multi-step variant of FGSM. The goal of the adversary is to solve the following type of problem:

$$\arg\max_{\|\delta\|_p \le \epsilon} \mathcal{L}(F_\theta(x+\delta), y). \tag{2}$$

In practice, the authors propose an iterative method to compute a solution:

$$x^{t+1} = \prod_{x \oplus \delta} (x^t + \alpha \text{sign}(\nabla_x J(x, y))), \tag{3}$$

where $x \oplus \delta$ is the *Minkowski* sum between $\{x\}$ and $\{\|\delta\|_p \leq \epsilon\}$, $\alpha$ is the gradient step size, and $\prod_S$ is the projection operator on *S*. Madry et al. showed that (the $l_\infty$ version of) PGD is equivalent to the basic iterative method (BIM), which is another important iterative attack. As for fast gradient methods, PGD can be implemented either with $p = \infty$ or $p = 2$. In this study, we use PGD to represent a variety of iterative attacks.

**Carlini & Wagner (CW) attack** (Carlini&Wagner et al., 2017) CW (2017) solves the following type of optimization problem to craft an $l_2$-norm adversarial example:

$$\begin{aligned} &\arg\min_{\eta \in \mathbb{R}^n} \|\eta\|_2 + c \cdot f(x + \eta) \\ &s.t. \quad x + \eta \in [0,1]^n \end{aligned}, \tag{4}$$

where *f* is the objective function, which is defined to drive the example *x* to be misclassified. *c* represents a suitably chosen constant. CW attack is an $l_2$ attack because the objective function aims to minimize the $l_2$ norm of the perturbation. CW can also be implemented as an $l_\infty$ attack, but this attack is not as effective as classical $l_\infty$ attacks.

### 2.1.2 Defense methods

Various defense mechanisms have also been employed to combat the threat from adversarial attacks. As discussed in Section 1, defense mechanisms can be divided into two categories: enhancing the robustness of the network and data preprocessing.

**Enhancing the robustness of the network** Adversarial training is a popular approach to defend against adversarial attack by adding the adversarial examples generated using one or more chosen attack models to the training set (Goodfellow et al., 2015; Araujo et al., 2019). This situation often results in increased robustness when the attack model used to generate the augmented training set is the same as that used by the attacker. Given that the defense mechanism designed to protect against one type of attack often offers poor performance against the other, some improvement methods, such as randomized adversarial training (Araujo et al., 2019) have also been gradually proposed. These defense methods show that data quality and training process both affect model quality, and data enhancement can improve the model capability to defend attacks. Defensive distillation (Papernot et al., 2015) is another typical method. Defensive distillation trains the classifier in two rounds using a variant of the distillation method. The approach reduces model sensitivity to input variations by decreasing the absolute value of the model's Jacobian matrix and introduces difficulties for attackers in generating adversarial examples. However, this approach fails to adequately protect against black-box attacks transferred from other networks.

**Data preprocessing** The second category includes a variety of data preprocessing methods. PixelDefend (Song et al., 2018), which was proposed by Song et al., can convert an image with interference into a clear image before inputting it to the classifier. Some researchers have perceived perceptual disturbances similar to noise and designed high-level representation guided denoiser (HGD) to eliminate such noise (Liao et al., 2018). Xiaojun Jia et al. proposed ComDefend, an end-to-end image compression model used to combat against adversarial examples (Jia et al., 2019). Simultaneously, detection-based defenses against adversarial examples from the steganalysis viewpoint (Liu et al., 2019) also show excellent performance advantages. In contrast to the first category, implementing this category is easier because retraining the neural network is no longer required. Although these methods have good performance, they mostly treat adversarial examples defense as a denoising process and may not fully explain the effect of adversarial examples on the network. Defending adversarial examples is difficult due to the complexity of building a theoretical model of adversarial example generation process.

### 2.2 Generative adversarial networks

Generative adversarial network (GANs) (Goodfellow et al., 2014), originally introduced by Goodfellow et al. (2014), is one of the most promising methods for unsupervised learning in complex distribution in recent years. GANs are a class of artificial intelligence algorithms implemented by a system of two neural networks contesting with each other in a zero-sum game framework. GANs have achieved impressive results in image generation (Denton et al., 2015), image editing (Zhu et al., 2016), and representation learning (Radford et al., 2016; Salimans et al., 2016). The quality and diversity of results from GANs have continued to improve, from generating simple digits and faces to synthesizing natural scene images, generating 1k photorealistic portraits, and producing 1000 object classes. Some modified versions of GAN, such as Condition-GAN (Mirza&Osindero, 2014), WGAN (Arjovsky et al., 2017), Cycle-GAN (Zhu et al., 2017) and SAGAN (Zhang et al., 2018), also demonstrate performance advantages in many scenarios. The key to the success of GAN is the idea of an adversarial loss that forces the generated images to be indistinguishable from real photos. This loss is particularly powerful for image generation tasks because this is exactly the objective that most computer graphics aim to optimize.

At present, GANs are also used in adversarial examples generation (Yu et al., 2018) and defense. For defense, Defense-GAN (Samangouei et al., 2018) trained a GAN as a denoiser to project samples onto the data manifold before classification. However, Athalye et al. (Athalye et al., 2018) found that these methods, including other input transformations, suffered from obfuscated gradient problem and can be circumvented by corresponding attacks. Boundary conditional GAN (Sun et al., 2019) is proposed to enhance the robustness of DNN against adversarial examples by generating boundary samples of a pretrained classifier. These methods all take advantage of GAN. The experiments in this paper are also compared with the two methods.

### 2.3 “Texture-Shape” perception of CNN

Convolutional neural networks (CNNs) are generally known to identify objects by learning increasingly complex shape representations. Some recent studies illustrate that image textures show an important role in CNN. Robert et al. (Geirhos et al., 2019) quantified two contradictory hypotheses by evaluating CNN and human observers using images with “texture-shape” conflicts. Their evaluation shows that the most important factor in the standard process of artificial intelligence for object recognition is not the shape but the texture. Then, the authors used the style transfer method to generate a new data set; thus, the shape of the object is the only remaining useful information. Research also suggests that CNN may tend to be more priority to identify images by shape if learning different textures of the same shape.

Recent research by the MIT team indicates that many applications of machine learning require models that are “human-aligned”, that is, they make decisions based on human-meaningful information regarding the input (Ilyas et al., 2019). Then, they reconverted robust training as a tool for enforcing human priors on the features learned by CNNs. The resulting robust feature representations turn out to be significantly more aligned with human perception (Engstrom et al., 2019).

However, the present research on the difference between neural network and human perception remains at the level of network characteristics. At present, no research links the separate “texture-shape" perception between networks and humans to the emergence of adversarial examples is available. The effects of this perception on the defense against adversarial examples have not been studied.

## 3. HAD-GAN model

We propose a new defense network architecture called human-perception auxiliary defense GAN (HAD-GAN). In this section, we will elaborate HAD-GAN and introduce the procedure of using our HAD-GAN to enhance the robustness of our pre-trained classifier against adversarial examples.

### 3.1 Motivation

Based on the above background and thinking, our model is mainly designed around the following three points:

- **Providing robust artificial intelligence.** The machine learning system is easily deceived by the change in input information. With the deployment of additional intelligence-based systems, solving the confrontation problem is of practical significance. Focusing on defense against adversarial examples, we design new defense models that aim to improve the robustness of the classification neural network.
- **Making the network learning process more like that of humans.** The samples encountered in human learning, including those with various disturbances, are diverse and ambiguous and are even combined with imagination. Some studies have found that network classification preferences can be adjusted through control input. Therefore, our defense model is designed based on human thinking and discriminant laws. By focusing on human perception, the model attempts to simulate the learning process of people.
- **Taking full advantage of GAN.** GAN is a new and successful deep model of generating models. Notably, GAN is similar to human beings in generating images through imagination. Therefore, when considering human perception preferences, we use the excellent image generation capabilities of GAN to design an auxiliary defense GAN model that is more in line with human experience.

### 3.2 Network architecture

As described in the picture, our network structure mainly comprises three sub-networks, including two mappings (**G**: Generator, **TTN**: Texture Transfer Network) and the Discriminator (**D**, contains the target Classifier **C**). The input to our network is the clean image in the dataset. Every generated sample has a corresponding class label *c* in addition to the noise *z*. **G** uses both to generate images: $(c,z) \rightarrow X_{fake}$. **TTN** uses the input image *x* and the texture dataset to generate real images for **D**: $(x,t_1,t_2,\cdots,t_N) \rightarrow (s_0,s_1,s_2,\cdots,s_N) \rightarrow X_{real}$. Then the output images of the two mappings $X_{fake}$ and $X_{real}$ are entered into a CNN containing the target classifier **C**: $X \rightarrow (c_1,c_2,\cdots,\ c_n)$.

Notably, we linked a fully connected discriminant network in parallel at the penultimate level of the target classifier. Thus, the target classifier and the discriminant layer together form an adversarial discriminator **D** in GAN, in which **D** aims to provide a probability distribution over sources and class labels. The training mechanism is also inspired by the ACGAN (Odena et al., 2017). In this framework, we only need to modify some parameters based on the image characteristics, and then defense training for almost any classifier can be achieved.

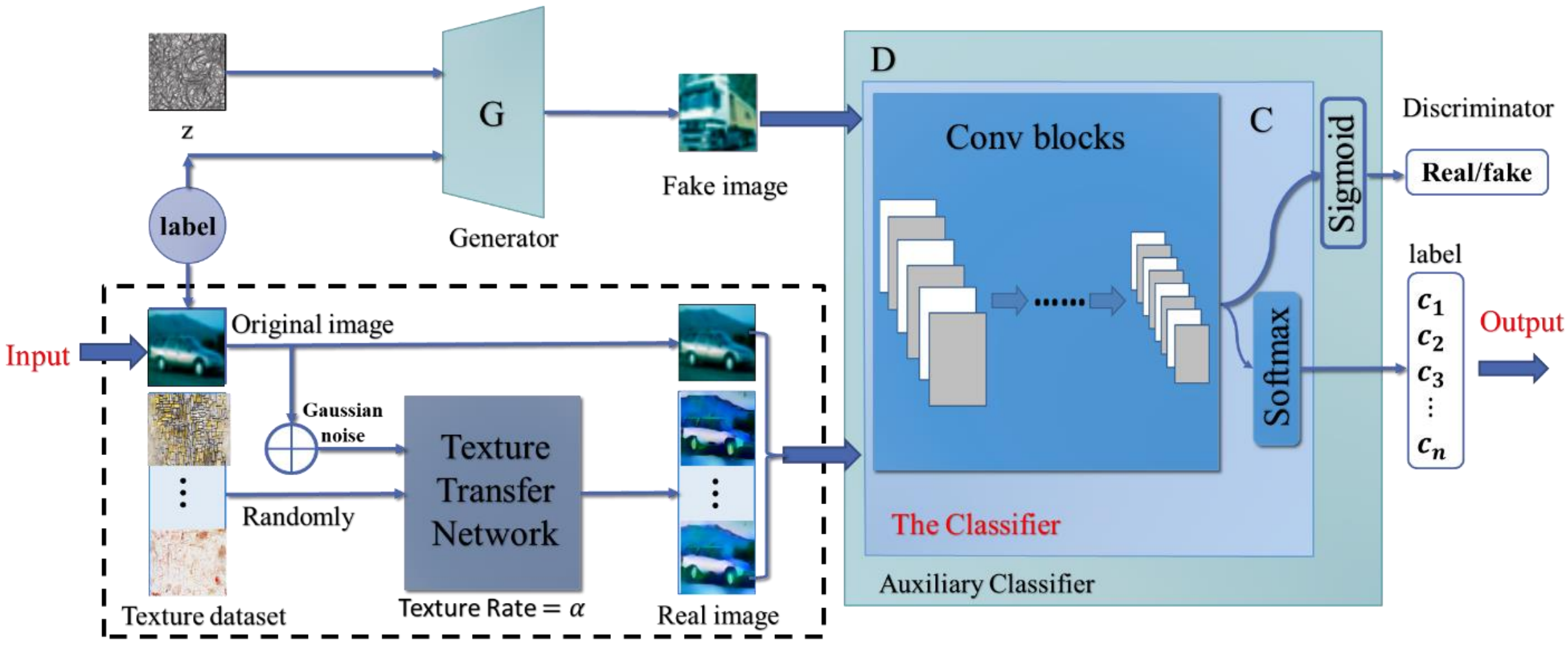

Figure 1. Model structure of HAD-GAN

### 3.3 Formulation

In this section, we present a set of definitions and main loss functions to help formally describe our framework, theoretical model and algorithm. We develop our framework with these loss functions to enable our network to learn weight parameters of **G** and **D** given training samples $\{x_i\}$. The Classifier becomes robust through the training on extending texture images. This section is mainly used to formulate each module and elements to facilitate the establishment of the latter model. The input to our model is the real image $X_{\text{real}}$ and its label $\{c\}$ in the database.

#### 3.3.1 Texture Transfer model

Our **TTN** uses an original image $x$ and a series of selected texture images as inputs and synthesizes output images corresponding to the textures that recombine the shape of the former and imitates the texture of the latter.

From the input batch $x \in \mathrm{R}^{N\times C\times H\times W}$, we strip every single $x$ of its original texture and replace it with the style texture of a randomly selected painting by adaptive instance normalization (AdaIN) style transfer (Huang& Belongie, 2017) with a texture coefficient of $\alpha$. We obtain the style images from Kaggle's Painter by Numbers dataset due to its large style variety and size.

AdaIN is a simple extension to instance normalization (IN), which we call adaptive instance normalization. AdaIN receives a content input x and a style input y and simply aligns the channel wise mean and variance of $x$ to match those of $y$. Unlike batch normalization, IN, or conditional instance normalization, AdaIN has no learnable affine parameters. Instead, AdaIN adaptively computes the affine parameters from the style input as follows:

$$\mathrm{AdaIN}(x, y) = \sigma(y)\left(\frac{x-\mu(x)}{\sigma(x)}\right) + \mu(y), \tag{5}$$

in which we simply scale the normalized content input with $\sigma(y)$ and shift it with $\mu(y)$. Similar to IN, these statistics are computed across spatial locations. $\sigma(x)$ and $\mu(x)$ are computed across spatial dimensions independently for each channel and each sample. Figure 2 shows an overview of our Texture Transfer Network based on the proposed AdaIN layer.

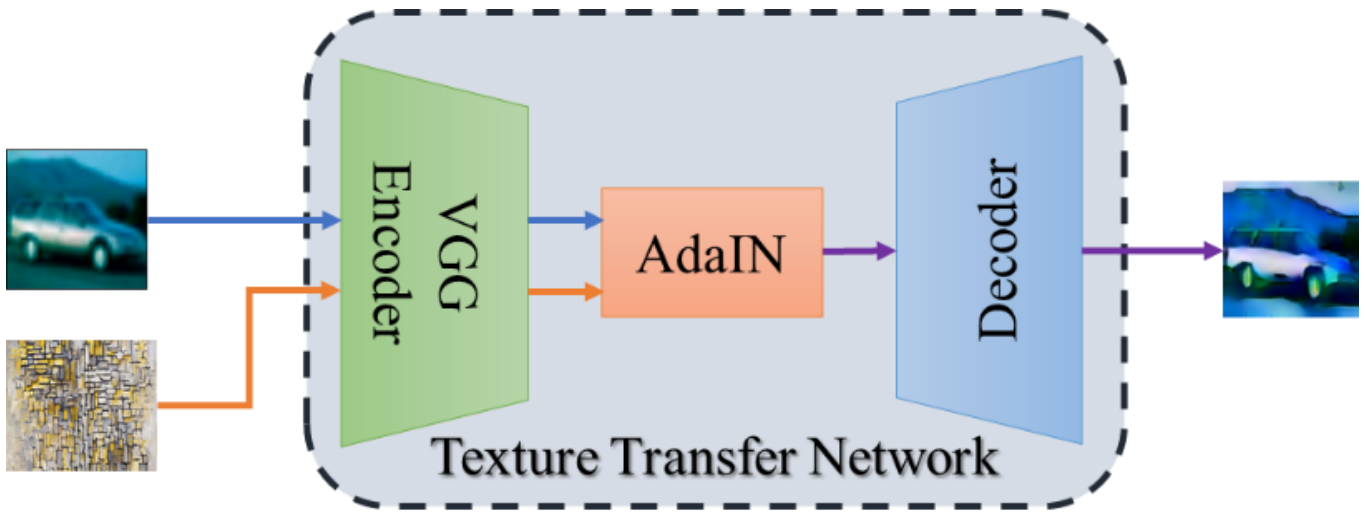

Figure 2. Model structure of TTN

**TTN** uses the first few layers of a fixed VGG-19 network to encode the content and style images. An AdaIN layer is used to perform style transfer in the feature space. A decoder is learned to invert the AdaIN output to the image spaces.

$$T(x,s)=g(t). \tag{6}$$

Feature maps

$$t=\text{AdaIN}(f(x),f(s)), \tag{7}$$

decoder *g* is trained to map t back to the image space. Generating the *T*(*x*, *s*),

$$T(x,s,\alpha)=g\left(\left(1-\alpha\right)f(x)+\alpha\text{AdaIN}\left(f(x),f(s)\right)\right), \tag{8}$$

$$t_{ij}=T(x_i,s_j,\alpha)=g\left(\left(1-\alpha\right)f(x_i)+\alpha\text{AdaIN}\left(f(x_i),f(s_j)\right)\right). \tag{9}$$

Instead of the commonly used feature responses of the content image, we use the AdaIN output *t* as the content target. $\alpha$ is texture weight parameter, which is a weight that controls the degree of texture. Given that AdaIN layer only transfers the mean and standard deviation of the style features, style loss only matches these statistics.

### 3.3.2 Adversarial Loss

We apply adversarial losses to mapping function **G** and its discriminator **D**. Given that they form the GAN structure, we can refer to the loss function of GAN.

$$\min_G \max_D V(D,G)=E_{x\sim p_{data}}[\log D(x)]+E_{z\sim p_z(z)}[\log(1-D(G(z)))]. \tag{10}$$

We facilitate the adversarial losses of ACGAN to improve our adversarial loss and train the target classification network in **D** simultaneously. The function becomes the following form with the class labels in addition to the model.

$$X_{\text{fake}}=G(c,z),D(x)=\{P(S\mid x),P(C\mid x)\}, \tag{11}$$

$$\min_G \max_D V(D,G)=E_{x\sim p_{data}}[\log D(x)]+E_{z\sim p_z(z),c\sim p(c)}[\log(1-D(G(c,z)))]. \tag{12}$$

The objective function is converted into the following two parts: the log-likelihood of the correct source, $L_S$, and the log-likelihood of the correct class, $L_C$.

$$L_S=E[\log P(S=real\mid X_{real})]+E[\log P(S=fake\mid X_{fake})], \tag{13}$$

$$L_C=E[\log P(C=c\mid X_{real})]+E[\log P(C=c\mid X_{fake})]. \tag{14}$$

Then, **D** and **G** are trained to maximize $L_S+L_C$ and $L_C-L_S$, respectively. As the ACGAN learns a representation for noise *z* that is independent of the class label (Odena et al., 2017), the original generator **G** loss and discriminator **D** loss is as follows:

$$L_{\mathbf{G}} = L_C - L_S \,, \tag{15}$$

$$L_{\mathbf{D}} = L_S + L_C \,. \tag{16}$$

Intuitively, the predictive distribution by the classifier for the samples near the decision boundary is close to a uniform distribution due to the ambiguity of the class to which the boundary samples belong. Therefore, generating samples close to the boundary has been proven to be a possible way to improve the robustness of the classification model (Sun et al., 2019). An additional factor known as confidence loss (CL) (Lee et al., 2017) is leveraged in our model loss to improve the performance.

$$L_{CL} = \beta \mathbb{E}_{P(x)} \left[ KL\left( U(y) \| P_\theta(y \mid x) \right) \right], \tag{17}$$

where $\theta$ are parameters of the original classifier, which are fixed during the training of the model. $U$ is the uniform distribution and $\beta > 0$ is a penalty parameter. $KL$ stands for KL divergence. Then the full generator **G** loss is

$$L_{\mathbf{G}(CL)} = L_C - L_S + L_{CL} \,. \tag{18}$$

Finally, our full objective can be written as

$$\mathbf{G},\mathbf{D} = \arg \min_{\mathbf{G}} \max_{\mathbf{D}} L_{\text{GAN}} \left( \mathbf{G}(c,z), \mathbf{D}(\mathbf{C},d) \mid T(x,s) \right). \tag{19}$$

### 3.3.3 Algorithm Flow

**Algorithm 1** HAD-GAN.

**Input:** dataset *D*, training epochs *K*, batch size *m*, texture number *T*, texture coefficient $\alpha$, noise coefficient *n*

**for** $k = 1$ **to** $K$ **do**

  **for** random batch $\{\mathbf{x}_i, y_i\}_{i=1}^{m} \sim D$ **do**

    **add noise (**the mean of the gaussian noise is 0 and the variance of the gaussian noise is *z***):**

      $x_i' = x_i + n \,,\, i = 1, \cdots, m$

    **for** $j = 1$ **to** $T$ **do**

      $T(x_i', s_j, \alpha) \leftarrow \textbf{decoder}\left( (1-\alpha) f(x_i') + \alpha \text{AdaIN}\left( f(x_i'), f(s_j) \right) \right)$

    **end for**

    **get the discriminator:** $\mathbf{D} \leftarrow \{\mathbf{C}, d\}$

    **generative adversarial training** (updating model parameters of **G** & **D**):

      $\theta_{\mathbf{G}}, \theta_{\mathbf{D}}, \theta_{\mathbf{C}} \leftarrow \arg \min_{\mathbf{G}} \max_{\mathbf{D}} L_{\text{GAN}} \left( \mathbf{G}(c,z), \mathbf{D}(\mathbf{C},d) \mid T(x,s) \right)$

**Output:** model parameter $\theta_{\mathbf{C}}$

## 3.4 Defense Mechanism

The designed defense network does not assume a known attack method but simply leverages the generative power of GAN. The robust classification network can be trained on either a pre-trained classification network or a complete training of the untrained classifier. The texture transfer network in our defense model is built on a pre-trained network to ensure defense efficiency. After this processing, the main adjustment parameters in **TTN** are input noise power and texture weight, which effectively improve the training efficiency of the entire defense network. Notably, selecting the appropriate texture database based on different classification problems can achieve excellent experimental results.

We describe our defense mechanism as follows, and the corresponding flow chart of the process is shown in Figure 3.

1. The target classifier is added to our defense models based on the network architecture. The well-trained TTN and the texture dataset are obtained.
2. Appropriate noise and texture weight are set. The noise in our model is set to Gaussian noise.
3. HAD-GAN is trained with the losses in Eq. (19) to stabilize the generation result of GAN network generator and the classification accuracy rate of auxiliary classifier.
4. Finally, the defensive-trained target classifier is taken out to calculate the robustness of the classifier against various adversarial examples.

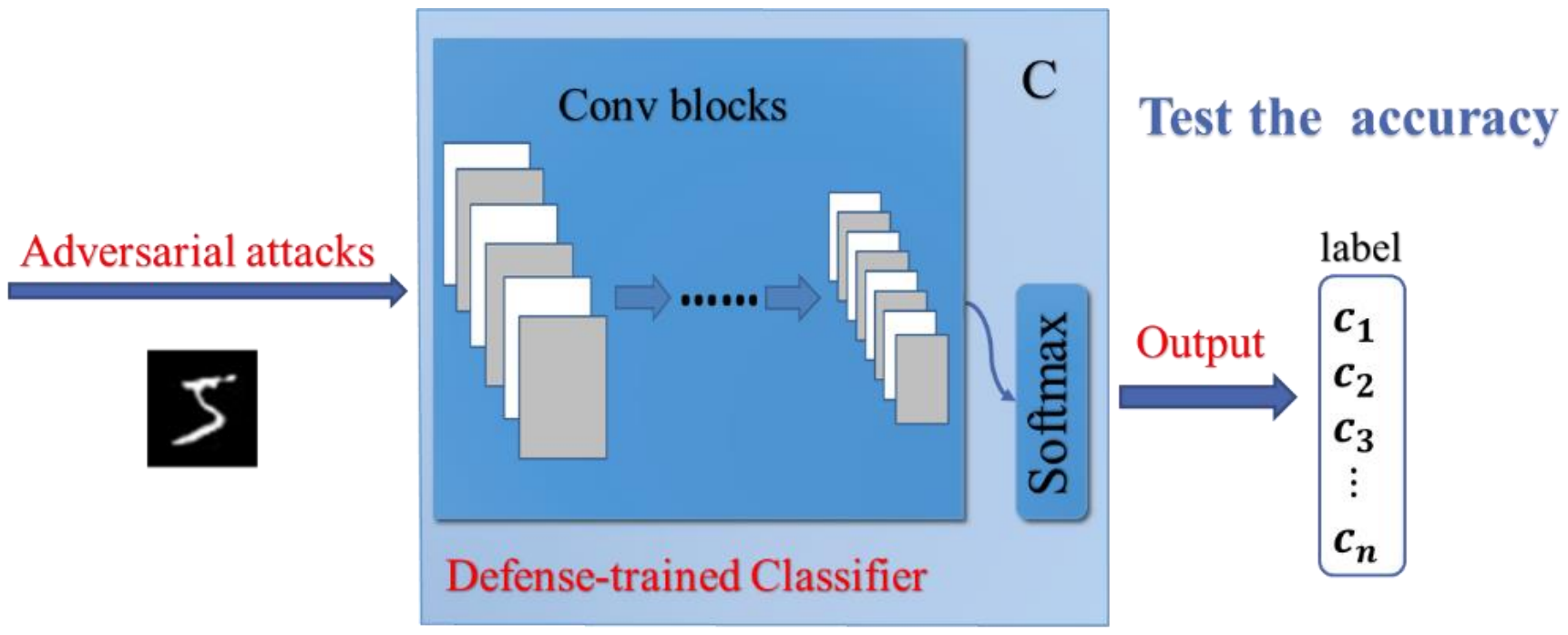

Figure 3. Defense mechanism

## 4. Experiments

We conducted experiments on the proposed model mainly on the MNIST, Fashion-MNIST, and CIFAR10 datasets. We quantitatively evaluate the network robustness on MNIST, Fashion-MNIST and CIFAR10 and compare it with other defensive methods to demonstrate the effectiveness of the proposed HAD-GAN. The experiment aims to prove that combining human discriminating preferences will improve the robustness of the network defense against the sample and explain the similarities and differences between the neural network and the human eye discriminating mechanism. Our method also successfully improves the accuracy of classification under attack conditions.

### 4.1 Experimental settings

We first selected the target classifier based on the dataset complexity. LeNet5 is assumed to be the target Classifier that our model attempts to defend for MNIST and Fashion-MNIST. We use ResNet18 instead for CIFAR10. As a model based on GAN, the generator is also crucial. We choose a basic generation network with four convolutional block layers for MNIST and Fashion-MNIST. The U-Net model is directly leveraged for CIFAR10. It's obvious that succeeding in an untargeted attack is easy, results in small perturbations, and transfers well to different models. Thus, we use untargeted adversarial examples to determine the performance of our model.

We test our defense model against untargeted attacks by FGSM, PGD, and CW. We conduct FGSM with different magnitude $\epsilon$ and PGD attack for 40 iterations with the gradient step size $\lambda$=0.01 on MNIST and Fashion-MNIST, with eight iterations on CIFAR10. Next, we conduct an $l_2$-norm attack with 2000 test samples. The accuracy of the original and defense-trained classifiers by HAD-GAN is computed to show the effectiveness.

### 4.2 Selection of hyperparameters

We initially construct our model by transferring the main focus of the neural network from the target texture to the shape in natural images. Therefore, our model cannot achieve excellent performance on MNIST and Fashion-MNIST at the beginning because images in monochrome cannot effectively perform texture conversion. We add Gaussian noise on input images to solve the problem, making texture generation easy. Furthermore, we find that images with small sizes cannot provide sufficient pixels to construct efficient textures. Thus, we adjusted the structure of the **TTN** model by adding two *Resize* modules. The small images are first enlarged to fit the texture image and then restored at the output end. For CIFAR10, additional noise is no longer necessary because the real image itself already easily implemented texture transformation. The adjusted **TTN** structure shown in Figure 4 demonstrates evident experimental results on MNIST, Fashion-MNIST and CIFAR10. The enlarged detail texture images are shown in Figure 5.

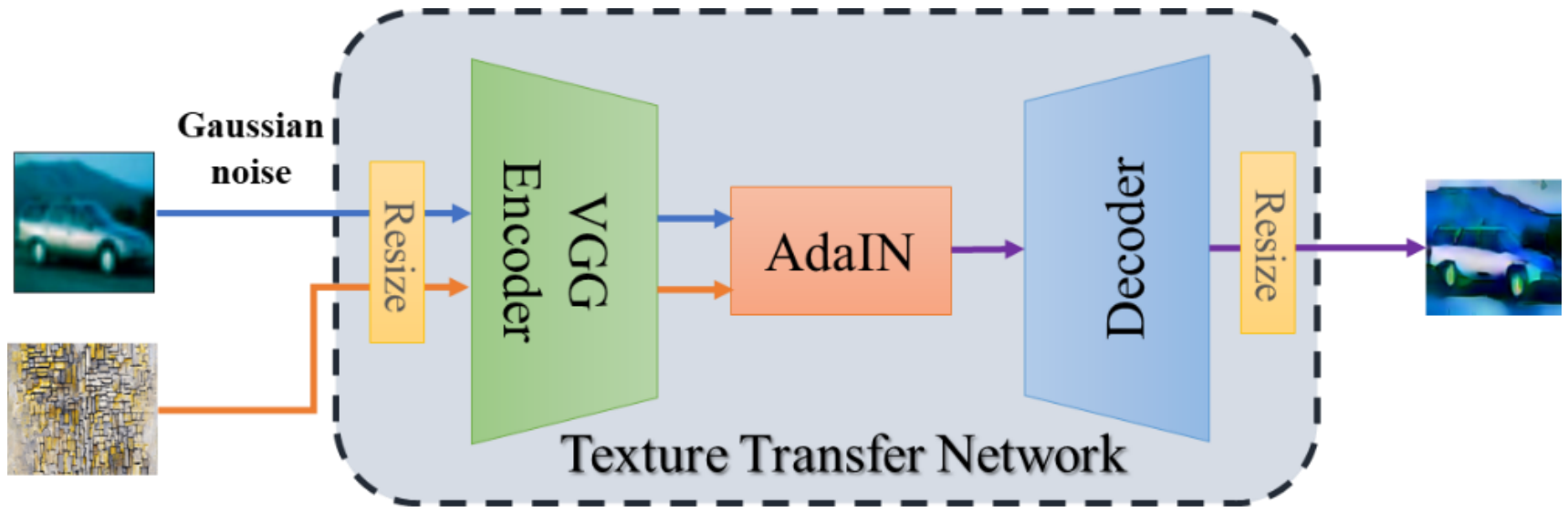

Figure 4. Improved TTN model structure

The above analysis indicates that the HAD-GAN must have improved performance on large nature images such as ImageNet. However, an advanced generator model of GAN is needed to generate realistic natural images. This goal may be beyond the scope of this study, and the relevant results of datasets, such as ImageNet, will be presented in future research.

Furthermore, the texture libraries corresponding to different types of datasets are different. For example, the images in MNIST and Fashion-MNIST are monochrome and the shapes are shown by white parts. We empirically found that the two can share a grayscale texture library. A separate color texture library is required for CIFAR-10, a. Experiments show that after proper selection of the texture library, adjusting the texture weight of **TTN** can make **D** achieve improved results. Overall, two hyperparameters in the neural networks must be determined before experiments. The first is the standard normal distribution Gaussian noise parameter $n$, and the second is the texture weight parameter $\alpha$. The value of $n$ and $\alpha$ depends on the performance of image auxiliary classification to improve the performance of the proposed method. Specifically, image compression discards part of the image information while retaining the main structural information of the image. The two hyperparameters are highly important in improving the model effect. Therefore, we first adjust the two parameters. For additional details, the main bases are as follows:

1. The infinite norm between the texture and the original image is calculated, that is $\|x_0 - t_i\|_\infty < \varepsilon$.
2. Compared with the standard training classifier, the classifier trained by the HAD-GAN model has a classification accuracy on clean samples of no less than 95%.
3. The changes in $L_G$ and $L_D$ are observed, and the training process of HAD-GAN tends to converge.

The parameter adjustment basis is complicated, but the experiment proves that satisfying any two of the three points can achieve ideal experimental results. In subsequent experiments, the corresponding parameter values for each dataset are shown in the following table:

Table 1

| Dataset | parameter | Value |
|---|---|---|
| MNIST | $n$ | 0.01 |
| | $\alpha$ | 0.3 |
| | $\epsilon$ | 0~0.3 |
| Fashion-MNIST | $n$ | 0.01 |
| | $\alpha$ | 0.2 |
| | $\epsilon$ | 0~0.2 |
| CIFAR10 | $n$ | 0 |
| | $\alpha$ | 0.03 |
| | $\epsilon$ | 0~0.05 |

Particularly, $\epsilon$ is the perturbation magnitude of the attack method for each dataset. Some enlarged textured images generated by **TTN** on the three datasets are shown in Figure 5. Additional textured images and images generated by **G** are shown in the **Appendices**.

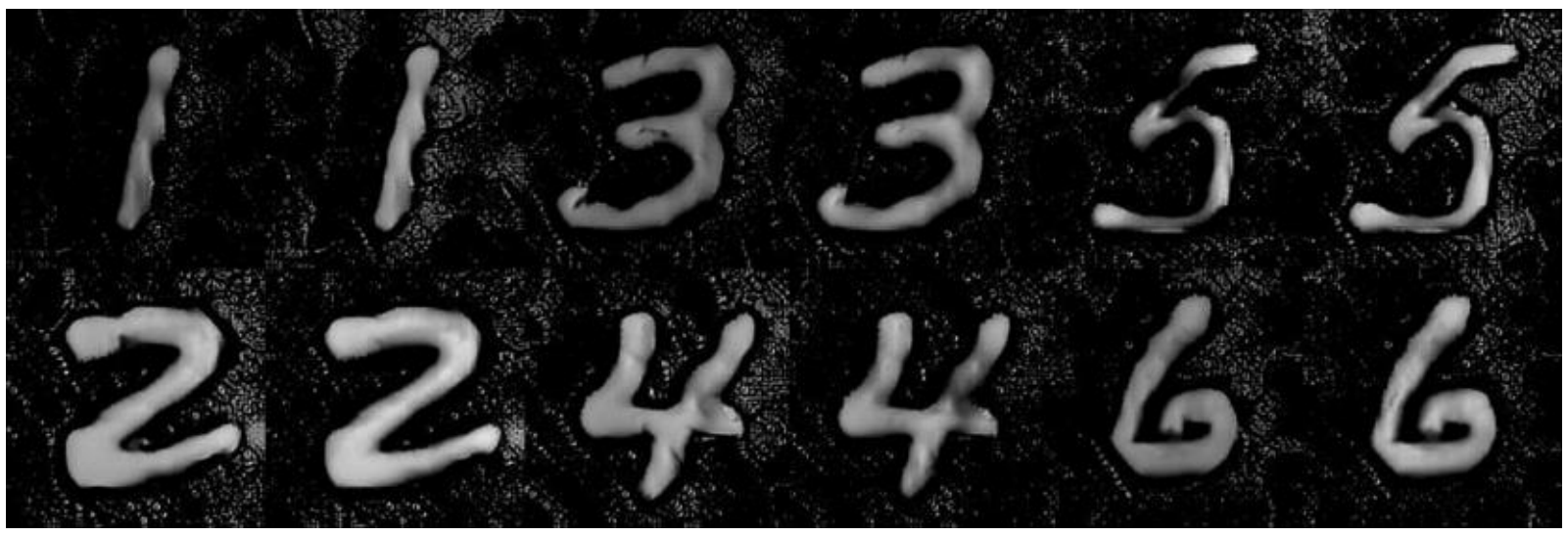

(a) textured images for MNIST

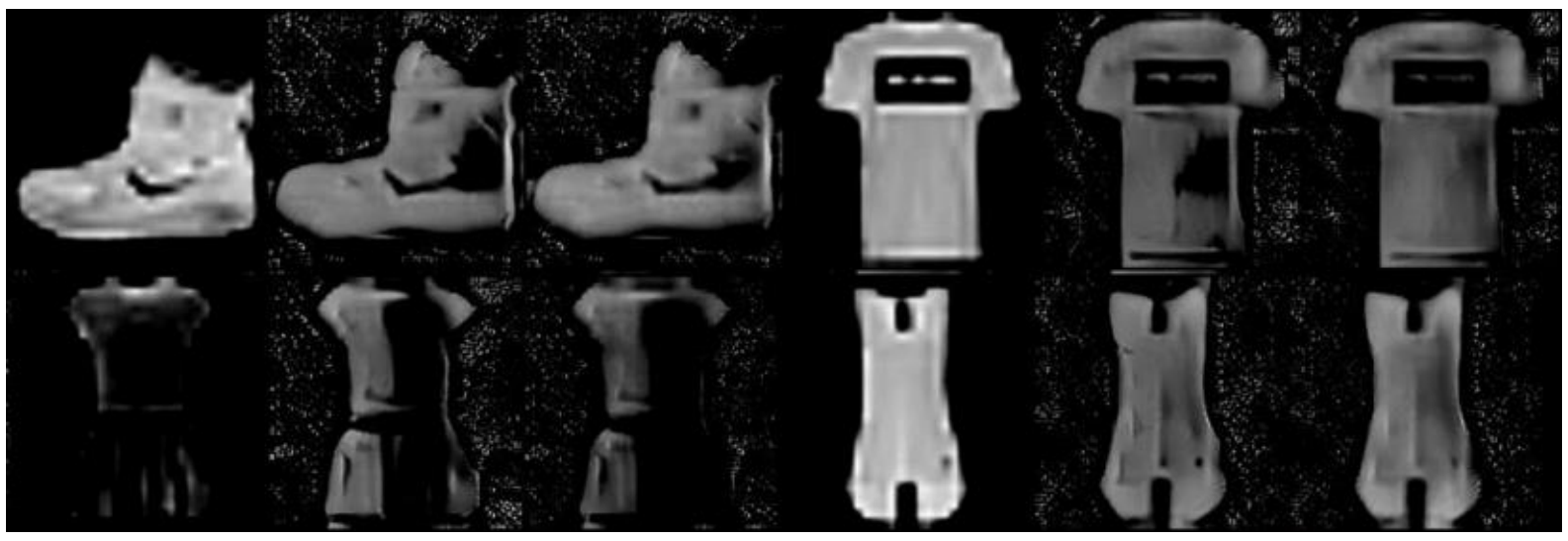

(b) textured images for Fashion-MNIST

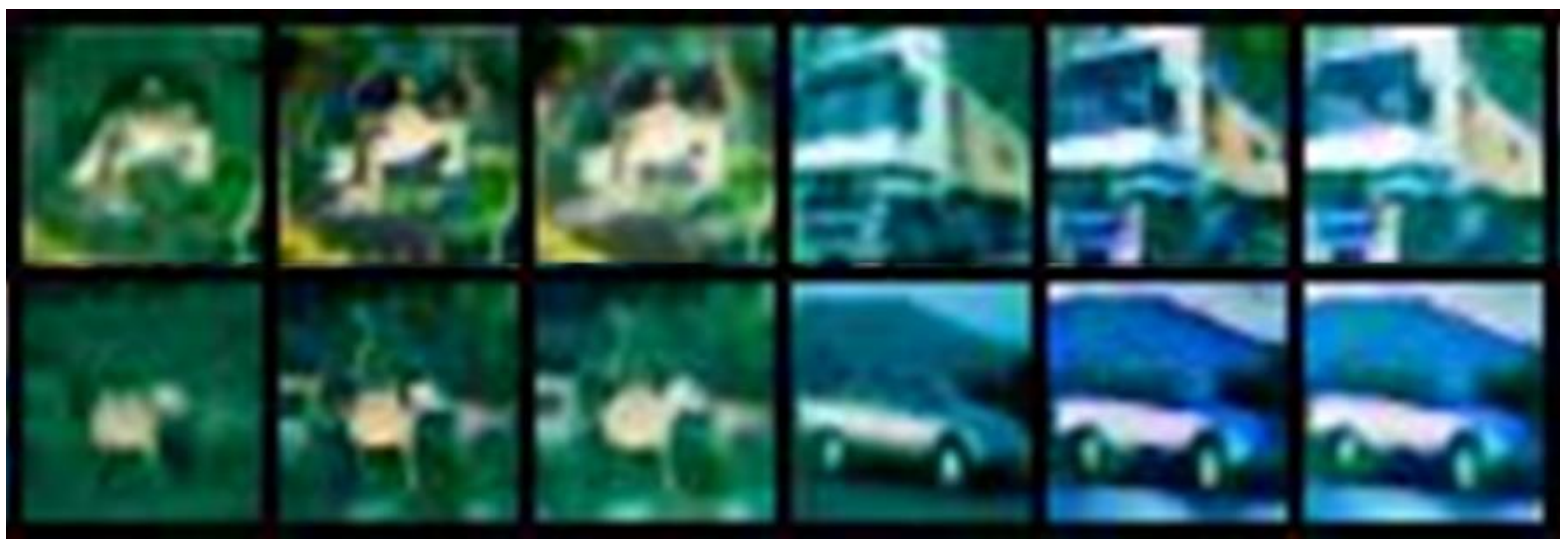

(c) textured images for CIFAR10

Figure 5. Enlarged textured images generated by **TTN**

Notably, for attack methods, such as FGSM, perturbation magnitude is set up to 0.3, as opposed to 0.4 or more because noises are evidently perceptible to human eyes when the attack is too strong (when $\epsilon > 0.4$). Thus, this part of the defensive experiment is not the focus of our research.

## 4.3 Result

### 4.3.1 Defensive performance testing

As previously mentioned, our model is mainly divided into the following three sub-networks: **TTN**, **G**, and **D**. We consider designing a comparison experiment, which compares the defensive performance of a module with and without it to verify the defense effect and analyze the performance of our model. For example, a round of experiments is performed without **TTN**, and then a set of comparative experiments is performed after the addition of **TTN**. The comparative tests of **TTN** are designed to verify our idea that adding human shape perception can enhance neural network defense performance against adversarial examples. Similarly, we investigate the effect of our defense model by computing the average accuracy of the partly- and fully-trained classifiers with our model which are tested on adversarial examples produced by different attack magnitude $\epsilon$ of FGSM and PGD.

Figure 6 respectively shows the experiment results on MNIST, Fashion-MNIST, and CIFAR10. The dotted lines (ND: No Defense) show the accuracy of image classification models tested on the adversarial images with no defense. ND is used as a baseline for defensive performance testing. CL is a Loss function added to the GAN to observe the performance improvement after enhancing the loss function. The red lines indicate the results without CL. The robustness of the model is significantly improved only after **TTN** addition (corresponding to human discrimination preference) to the training model. However, similar to the human eyes, the accuracy rate still decreases in the case of excessive attack perturbation, which also reflects the correlation between network design and human characteristics. We combine CL during the training process of GAN and continue the experiment to slow down the rate of accuracy decline. As shown by the blue line, optimizing the network parameters such as the loss function will also have a certain effect on the improvement of the defensive performance. This part of the experiment also shows that combining the advantages of the human eye and the network will be a feasible way to generate "1 + 1 > 2" and improve network performance.

In this respect, the fully-trained classifiers by our method increase accuracy on the FGSM strongest attack from 16.5% to 90.5%, the PGD strongest attack from 8.5% to 85.5% for MNIST; the FGSM strongest attack from 17.5% to 80.3%, the PGD strongest attack from 1.5% to 77.8% for Fashion-MNIST; the FGSM strongest attack from 29% to 78.6%, the PGD strongest attack from 5.2% to 76.5% for CIFAR10.

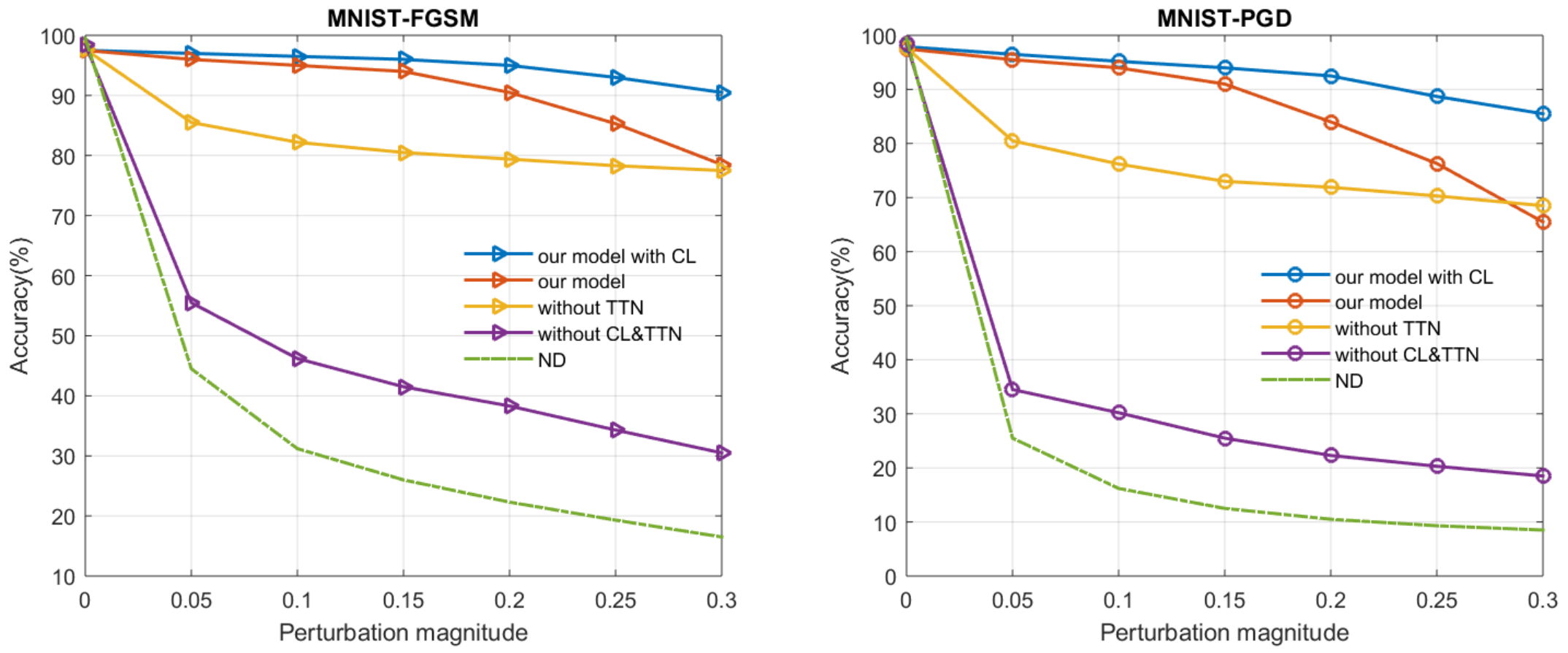

(a)

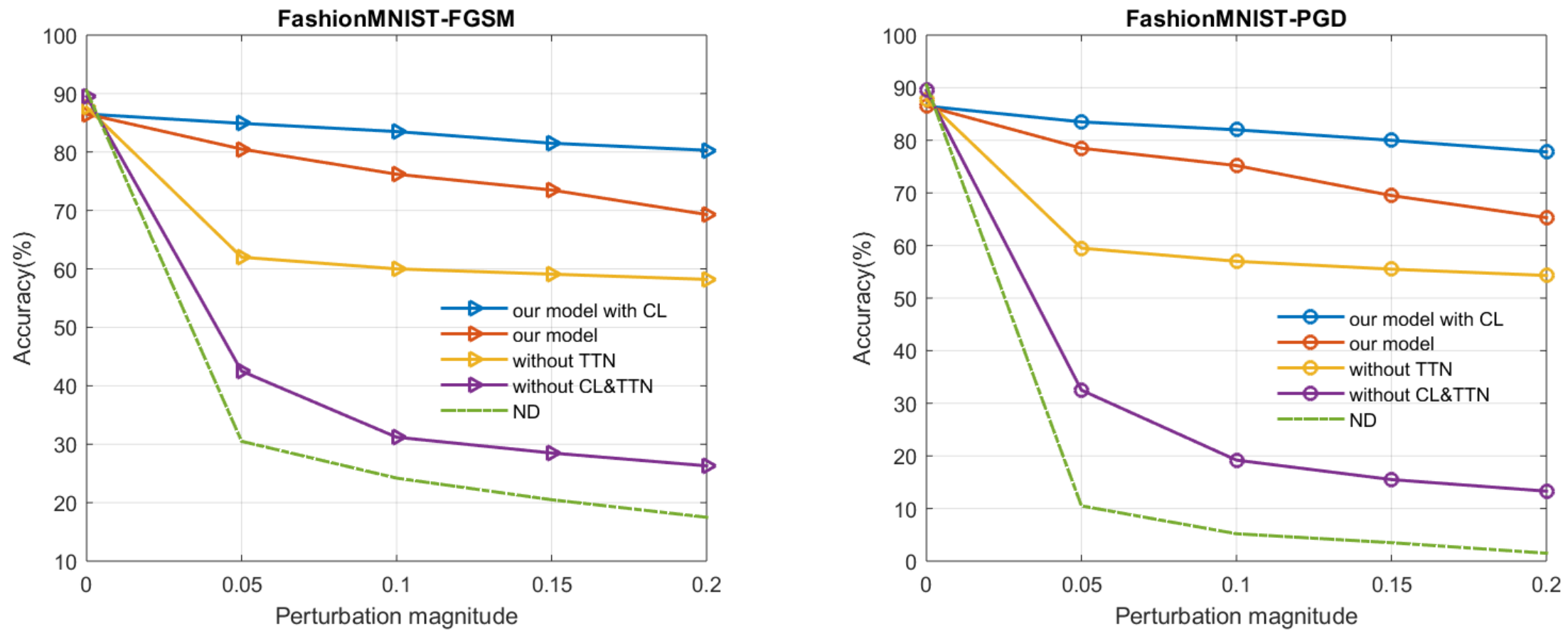

(b)

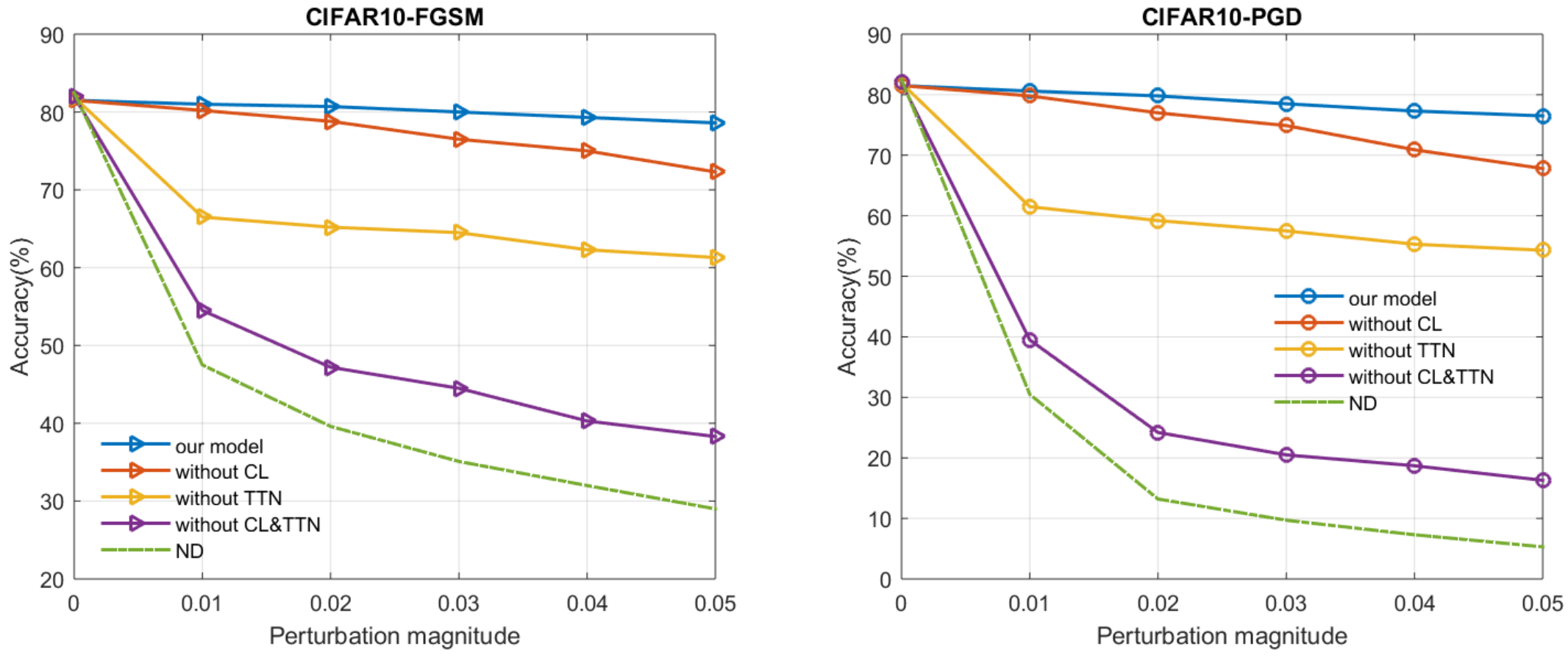

(c)

Figure 6. Average accuracy of our defense-trained classifiers tested on FGSM and PGD for different image datasets: (a) MNIST; (b) Fashion-MNIST; (c) CIFAR10

### 4.3.2 Comparisons with other defensive methods

We compare the proposed method with other conventional schemes under the same attack distance metric to measure the performance of our proposed method quantitatively. The results of the comparison on MNIST, Fashion-MNIST, and CIFAR10 image datasets are shown in Tables 2–4. Compared with these methods, our proposed method improves performance. In particular, our method achieves nearly 95% accuracy on the FGSM, PGD, and CW for MNIST; nearly 85% accuracy on the FGSM, PGD, and CW for Fashion-MNIST; nearly 80% accuracy on the FGSM, PGD, and CW for CIFAR10. The proposed method also achieves highly defensive accuracy against FGSM, PGD, and CW, and competitive accuracy compared with other defense methods, such as adversarial training and BCGAN. These improvements demonstrate the effectiveness of the proposed method. Compared with the classifier trained on normal methods, the clean images classification accuracy of the classifier trained on our model does not considerably decline. This result also shows that our model does not affect the normal classification of clean images by the classifier.

The core advantage of our method is that our network is trained on clean images rather than adversarial examples. Thus, we do not need to use attacking methods to generate adversarial examples, and our model is much smaller than many defense methods based on adversarial training. Furthermore, our proposed method does not depend on attacking methods and classifiers and can be combined with other defensive methods.

Table 2

| Defensive method | Clean | FGSM(0.2) | PGD(0.2) | CW |
|---|---|---|---|---|
| No Defense | 99.5 | 22.3 | 10.5 | 24.2 |
| FGSM Adv.Tr(0.2) | 99.4 | 88.9 | 53.2 | 65.3 |
| PGD Adv.Tr(0.2) | 99.5 | 87.6 | 89.6 | 79.2 |
| Distillation | 98.8 | 93.5 | 91.6 | 3.5 |
| Defense-GAN | 91.2 | 75.5 | 72.0 | 30.3 |
| BCGAN | 97.7 | 86.3 | 84.7 | 78.5 |
| Our Model | 97.5 | **95.0** | **92.5** | **89.3** |

Table 3

| Defensive method | Clean | FGSM(0.1) | PGD(0.1) | CW |
|---|---|---|---|---|
| No Defense | 90.5 | 24.2 | 5.2 | 0 |
| FGSM Adv.Tr(0.1) | 90.2 | 79.0 | 16.5 | 55.5 |
| PGD Adv.Tr(0.1) | 89.5 | 80.7 | 81.5 | 75.6 |
| Distillation | 88.5 | 76.5 | 75.0 | 0.0 |
| Defense-GAN | 80.2 | 65.5 | 43.0 | 37.5 |
| BCGAN | 85.4 | 72.8 | 71.5 | 68.9 |
| **Our Model** | 86.5 | **83.5** | **82.0** | **80.5** |

Table 4

| Defensive method | Clean | FGSM(0.03) | PGD(0.03) | CW |
|---|---|---|---|---|
| No Defense | 82.5 | 35.1 | 9.7 | 3.2 |
| FGSM Adv.Tr(0.03) | 82.0 | 58.9 | 33.2 | 20.9 |
| PGD Adv.Tr(0.03) | 79.8 | 57.3 | 30.6 | 28.4 |
| Distillation | 85.8 | 71.2 | 70.1 | 0.0 |
| Defense-GAN | 71.2 | 55.5 | 42.0 | 30.3 |
| BCGAN | 80.1 | 67.5 | 66.5 | 65.0 |
| **Our Model** | 81.5 | **80.0** | **78.5** | **75.0** |

## 5. Conclusion

In this paper, we propose a novel defense model to defend adversarial examples. HAD-GAN

can be trained on clean images without known attack methods. For the first time, HAD-GAN combined the shape-texture discrimination preferences of humans and neural networks in the defense against adversarial examples. Experimental results illustrate the effectiveness of our model. More importantly, our work demonstrates that adding human features to the network can improve the robustness of the network against attacks.

Researchers must always make a choice. Should the model make an "accurate" decision, or a "human" decision? The research in this paper further demonstrates that if we want to obtain a method that is meaningful to humans and faithful to the model, we must provide necessary interventions during the training process. At present, image recognition technology has been widely used in daily life, and we need to find a certain balance between the two choices.

**Acknowledgment**

This work was funded by the National Key R&D Program of China under grant 2017YFB1002502 and National Natural Science Foundation of China (No. 61701089 and No.61601518).

**Appendices**

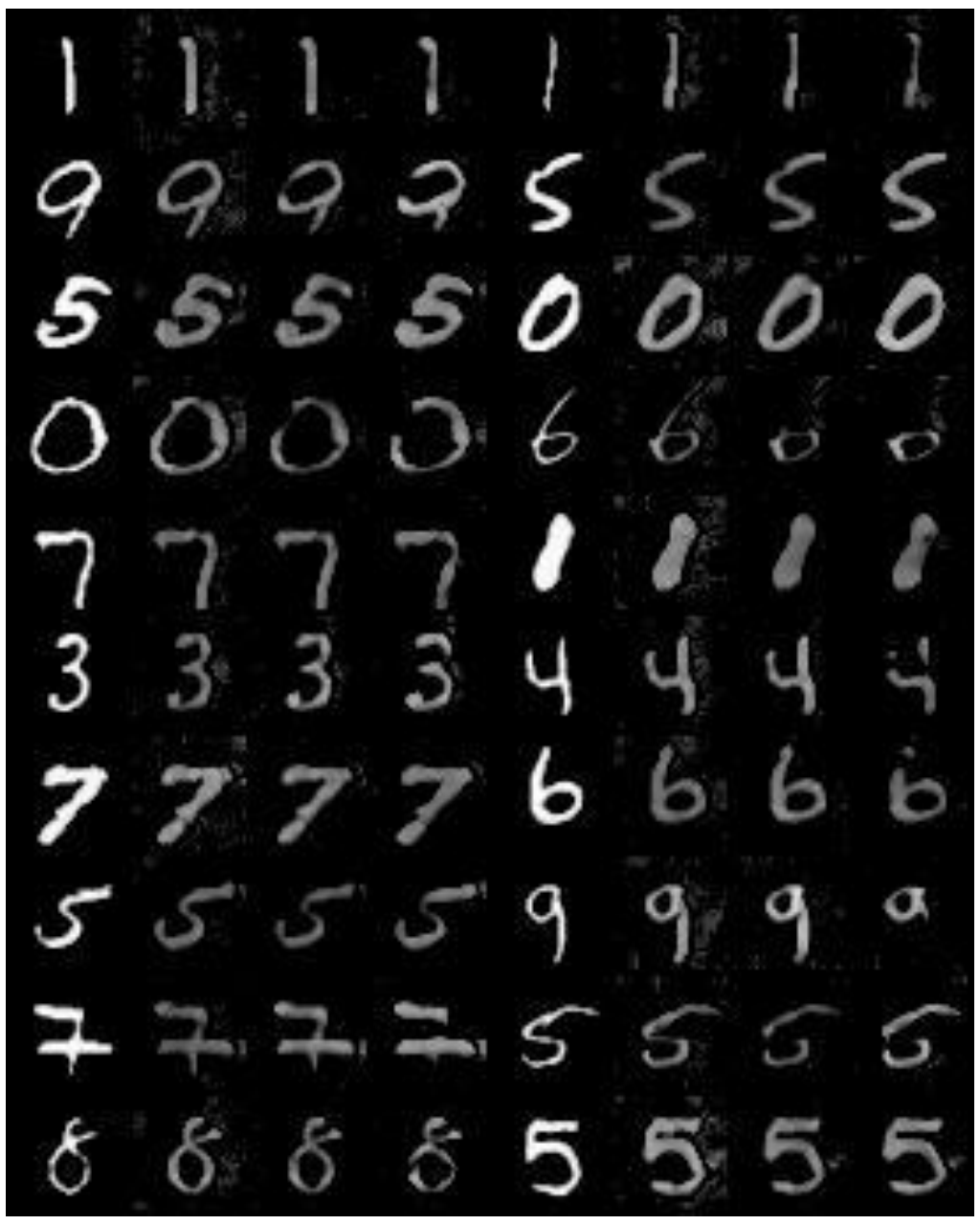

(a)

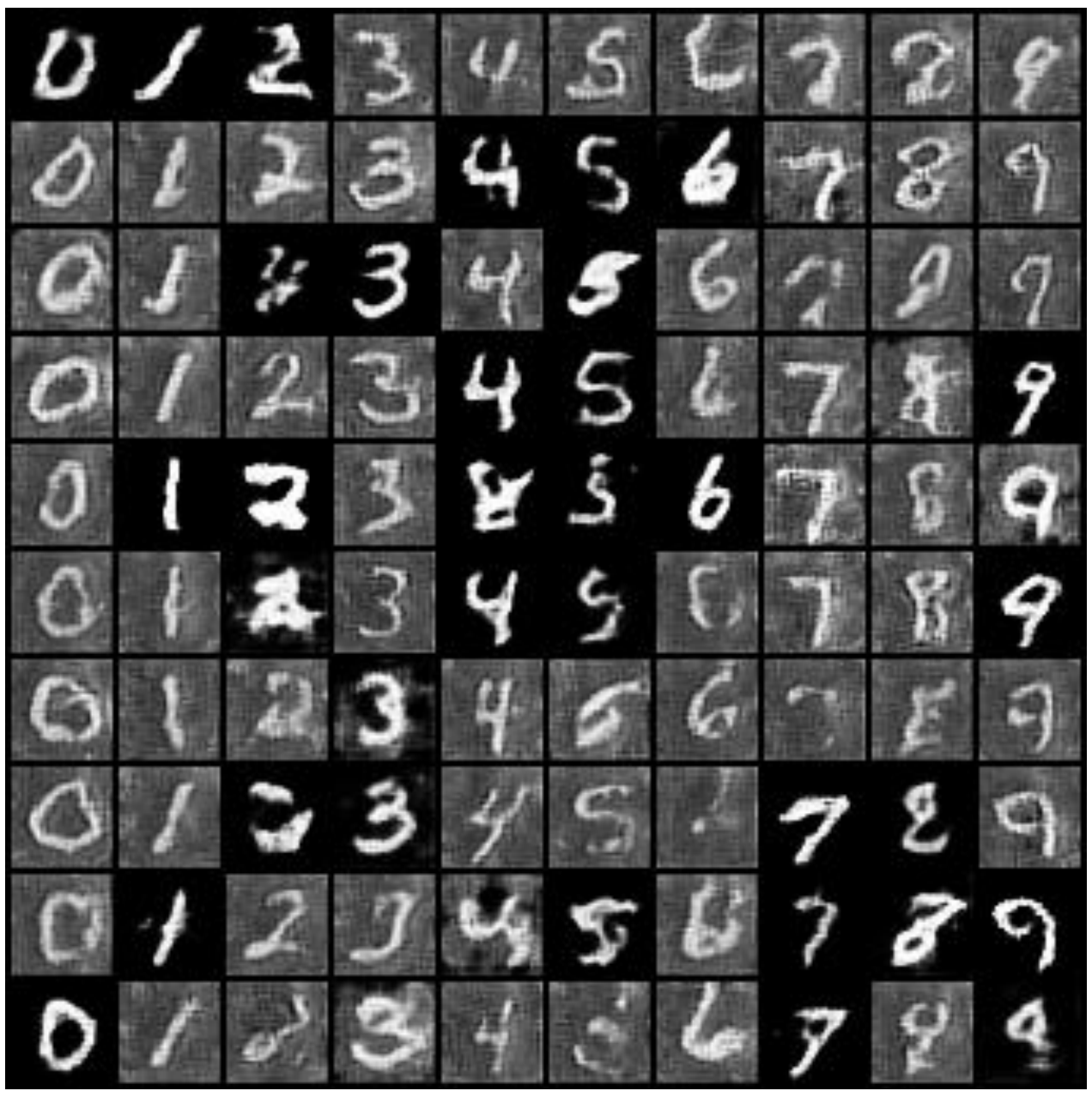

(b)

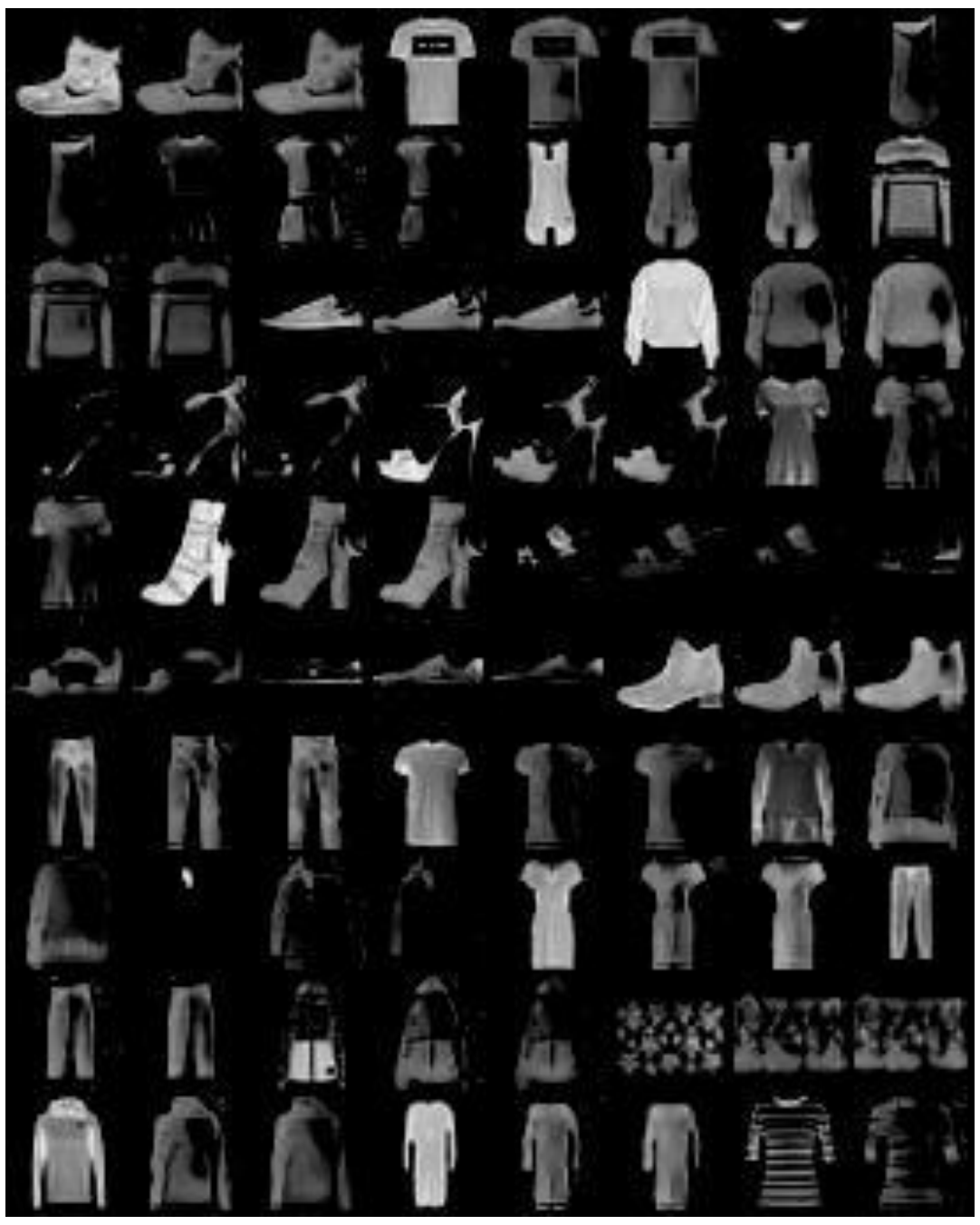

(c)

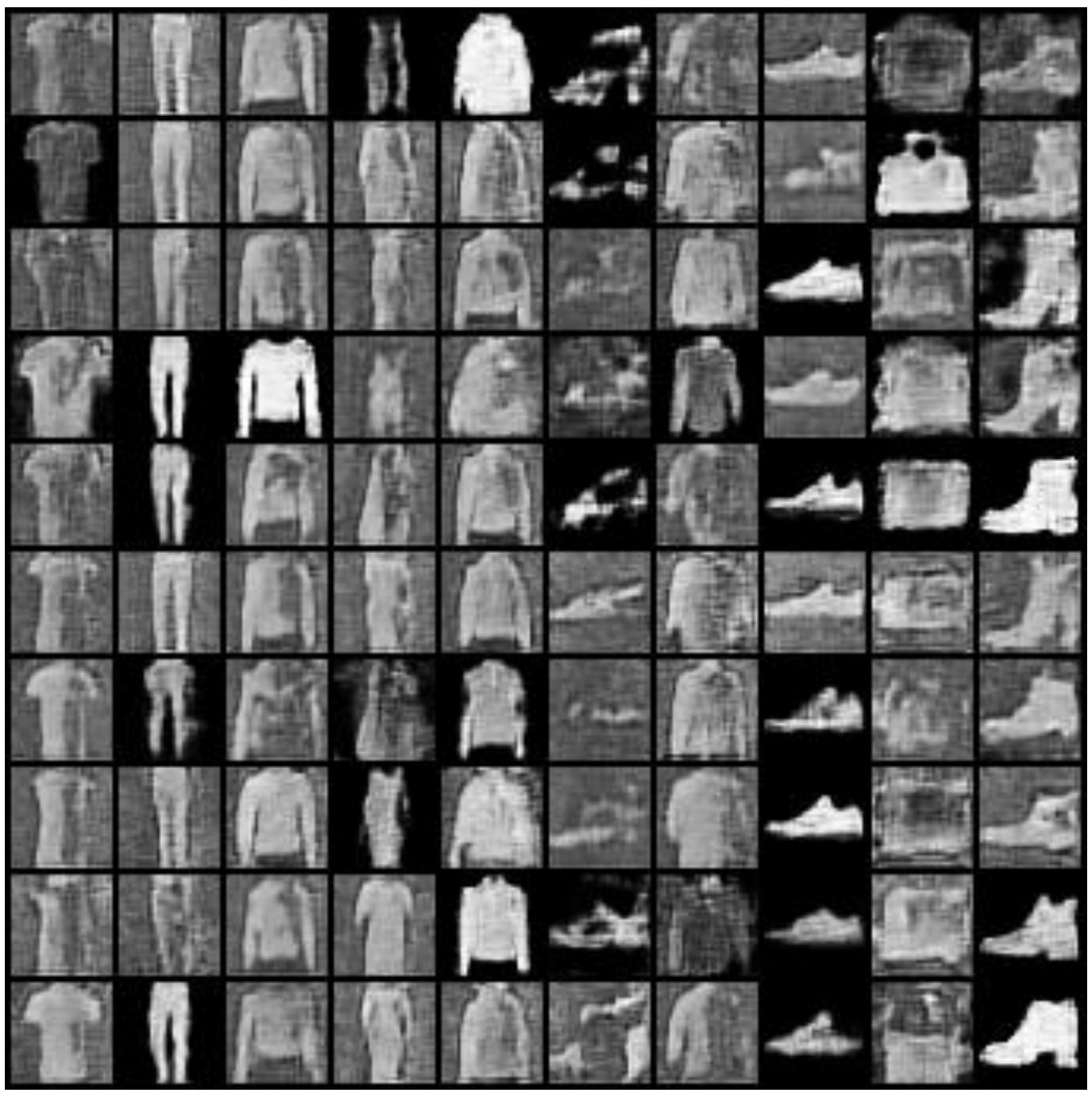

(d)

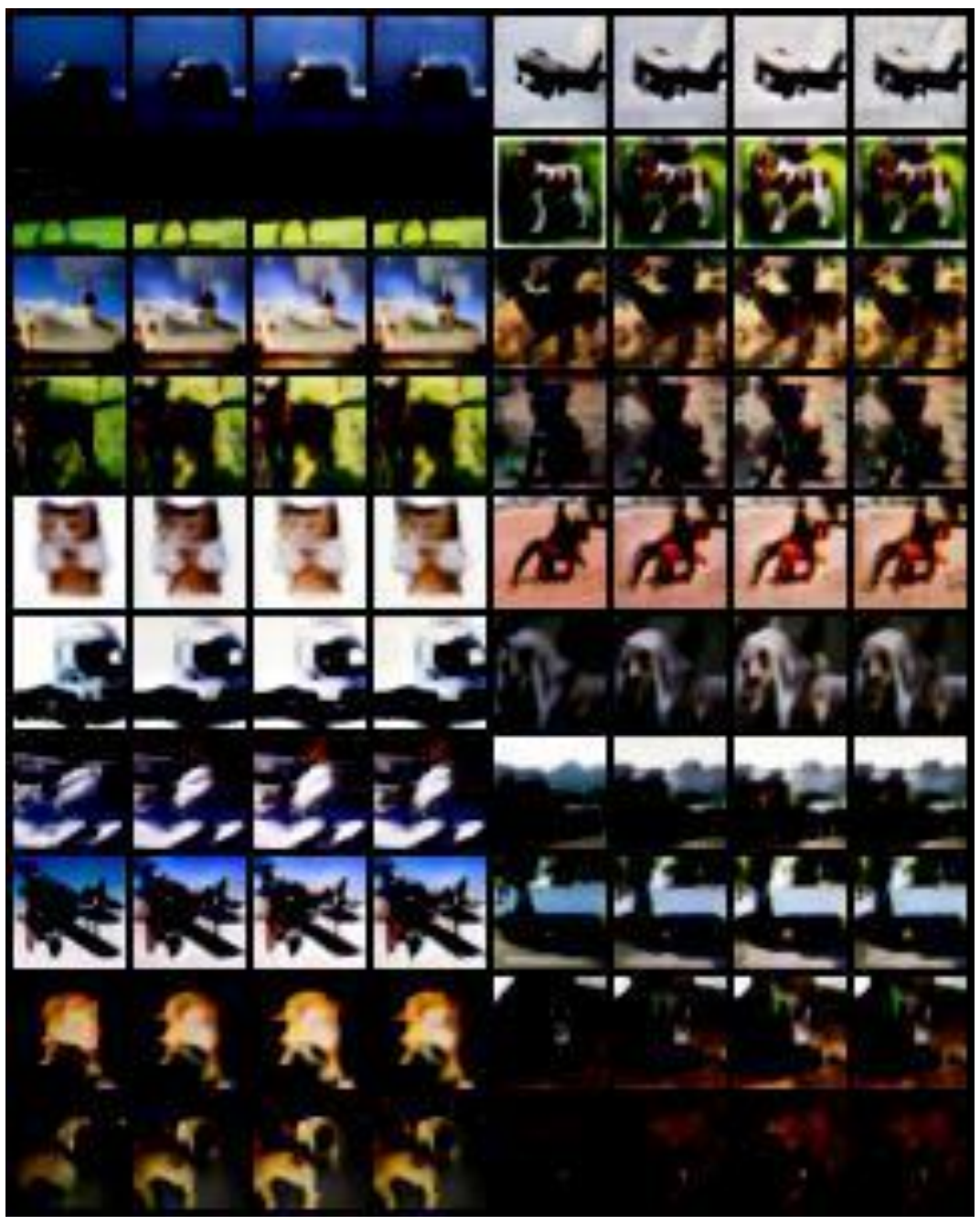

(e)

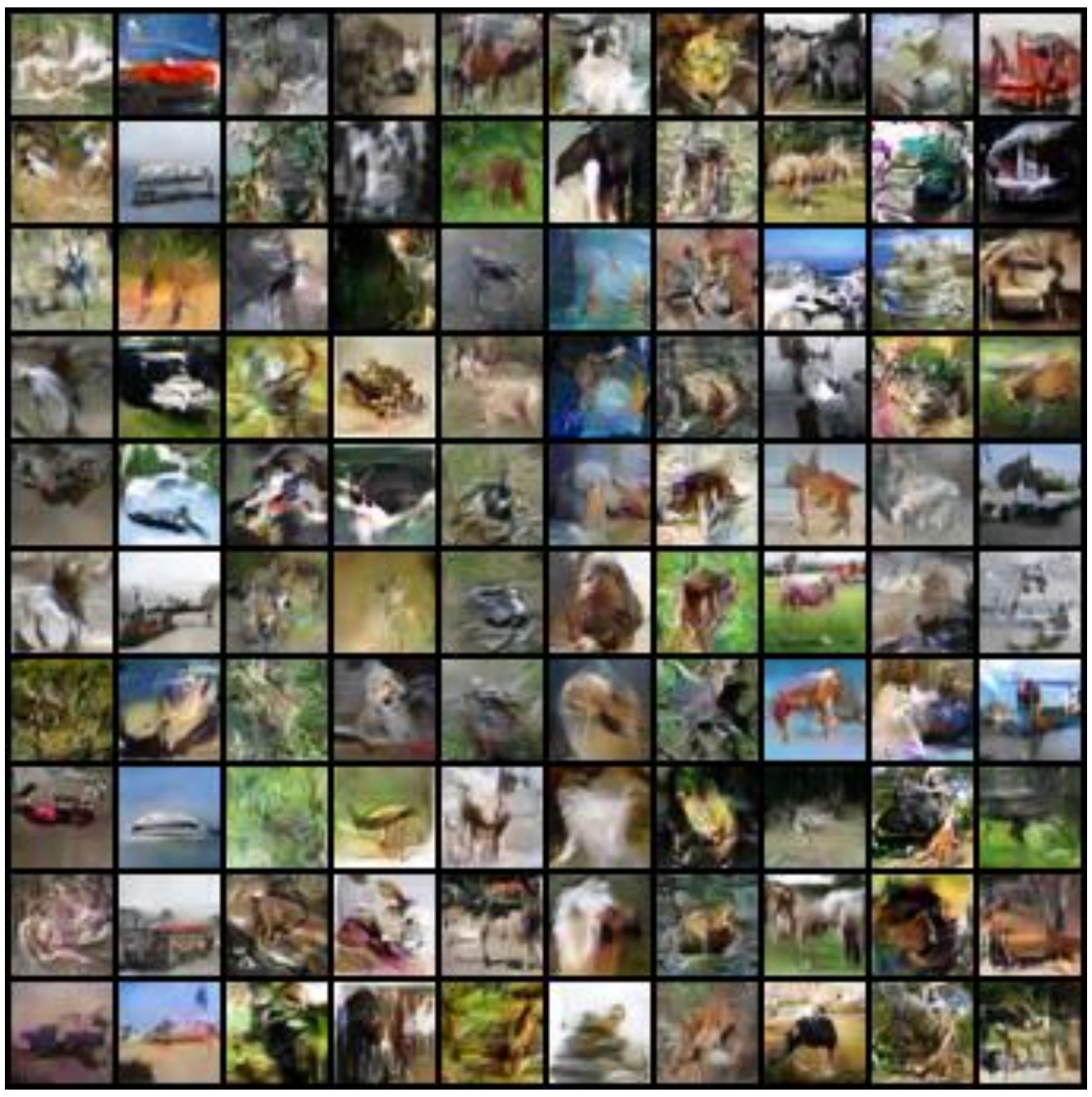

(f)

Figure 7: Examples of textured images generated by **TTN** on the three datasets and images generated by **G**. (a) Textured images for MNIST. Four consecutive pictures are a group; the first one is the original image, and the last three are the output images corresponding to different texture data. (b) Generated images for MNIST. Each column corresponds to a category. (c) Textured images for Fashion-MNIST. Three consecutive pictures are a group; the first one is the original image, and the last two are the output images corresponding to different texture data. (d) Generated images for Fashion-MNIST. Each column corresponds to a category. (e) Textured images for CIFAR10. Four consecutive pictures are a group; the first one is the original image, and the last three are the output images corresponding to different texture data.

Two interesting phenomena are worth pointing out. First, as shown in picture (c), distinguishing some original figures into categories by humans is difficult; for example, in the third set of pictures, the original picture can only see a collar-like area. However, after processing by the TTN network, pixels that are invisible to the human eye on the original image are displayed. After comparison, the texture image categories of these outputs are correct. Second, the generated images are randomly divided into two types: clear and blurred background. This division indicates that the networks trained by our model not only process clear images but also process and classify blurred images with noise interference.